\documentclass[10pt,twoside]{article}

\usepackage{times}
\usepackage[utf8]{inputenc}
\usepackage[T1]{fontenc}
\usepackage{graphicx}

\usepackage{taln2018}
\usepackage[frenchb]{babel}

\title{Système de traduction automatique statistique Anglais-Arabe}

\author{Marwa Hadj Salah\up{1, 2}\quad Didier Schwab\up{1} Hervé Blanchon\up{1} Mounir Zrigui\up{2}\\
  {\small
    (1) LIG-GETALP, Univ. Grenoble Alpes, France \\
    \texttt{Prénom.Nom@univ-grenoble-alpes.fr \\}
    (2) LaTICE, Tunis, 1008, Tunisie \\ 
    \texttt{Prénom.Nom@fsm.rnu.tn} 
}}

\begin{document}
\maketitle

\section{Introduction}

La traduction automatique (TA) est le processus qui consiste à traduire un texte rédigé dans une langue source vers un texte dans une langue cible. Dans cet article, nous présentons notre système de traduction automatique statistique anglais-arabe. Dans un premier temps, nous présentons le processus général pour mettre en place un système de traduction automatique statistique, ensuite nous décrivons les outils ainsi que les différents corpus que nous avons utilisés pour construire notre système de TA.

\section{Traduction automatique}

\subsection{Traduction automatique statistique }
\begin{center}
\begin{figure}[ht!]
\centering
\includegraphics[scale=0.5]{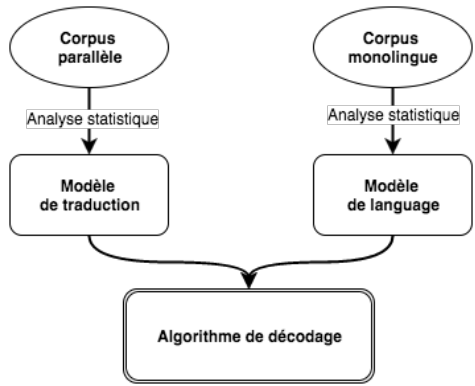}
\caption{Processus de la traduction automatique statistique}
\label{FigureTA}
\end{figure}
\end{center}

La traduction automatique statistique (TAS) est une approche très utilisée dans la TA et qui se base sur l'apprentissage de modèles statistiques à partir de corpus parallèles. En effet, comme il est montré dans la figure \ref{FigureTA}, la traduction automatique statistique se base essentiellement sur: Un modèle de langage (ML), un modèle de traduction (MT) et un décodeur.

\subsubsection{Modèle de langage}

Parmi les modèles de langages utilisés dans les systèmes de TAS les principaux sont le modèle n-gramme, le modèle Cache  \cite{kuhn1990cache} et le modèle Trigger \cite{lau1993trigger}. Le modèle Cache repose sur les dépendances des mots non contigus. Quant à lui, le modèle Trigger consiste à déterminer le couple de mots (X, Y) où la présence de X dans l’historique déclenche l’apparition de Y.

Toutefois, le modèle n-gramme (1$\leq$n$\leq$5) reste le plus utilisé dans les systèmes de traduction actuels et plus précisément le modèle trigramme ( -gramme pour le traitement des langues européennes. En effet, le modèle n-gramme permet d’estimer la vraisemblance d’une suite de mots en lui attribuant une probabilité.

Soit $\textit{t} = w_{1}w_{2} . . . w_{k}$
 une séquence de k mots dans une langue donnée et n la taille maximale
des n-gramme (1$\leq$n$\leq$5, la formule de p(t est exprimée en : 
\begin{equation}
 P(t)=\prod_{i=1}^{k} (w_{i}|w_{i-1}w_{i-2} ... w_{i-n+1})
\end{equation}

\subsubsection{Modèle de traduction à base de segments}  
 Pour construire un modèle de traduction à base de segments \cite{och2003systematic} , il est nécessaire de passer par trois étapes indispensables: 
\begin{itemize}
\item 	Segmentation de la phrase en séquences de mots
\item Traduction des séquences de mots en se fondant sur la table de traduction
\item Ré-ordonnancement des séquences de mots à l’aide d’un modèle de distorsion 
\end{itemize}

 \subsubsection{Décodeur}

Moses  \cite{koehn2007moses} est une boite à outils disponible sous licence libre GPL, basée sur des approches statistiques  de  la traduction automatique. En effet, Moses nous permet de développer et manipuler un système de traduction selon nos besoins grâce à ses nombreuses caractéristiques, telle que la production du modèle de traduction et le modèle de réordonnance à partir des corpus volumineux. \\
Parmi les principaux modules du Moses, on trouve  : 
\begin{itemize}
\item \textbf{Train} : permet de construire des modèles de traduction ainsi que des modèles de réordonnance.
\item \textbf{Mert}  : permet d'ajuster les poids des différents modèles afin d’optimiser et maximiser la qualité de traduction en utilisant les données de développement (DEV) . 
\item \textbf{Décodage} : ce module contient des scripts et des excusables permettant de trouver la traduction la plus probable d’une phrase source en consultant les modèles du module Train. 
\end{itemize}

\subsection{Outils}
 \subsubsection{Le décodeur Moses}

Moses \cite{koehn2007moses} est une boite à outils disponible sous licence libre GPL, basée sur des approches statistiques  de  la traduction automatique. En effet, Moses nous permet de développer et manipuler un système de traduction selon nos besoins grâce à ses nombreuses caractéristiques, telle que la production du modèle de traduction et le modèle de réordonnance à partir des corpus volumineux. \\
Parmi les principaux modules du Moses, on trouve  : 
\begin{itemize}
\item \textbf{Train} : permet de construire des modèles de traduction ainsi que des modèles de réordonnance.
\item \textbf{Mert}  : permet d'ajuster les poids des différents modèles afin d’optimiser et maximiser la qualité de traduction en utilisant les données de développement (DEV) . 
\item \textbf{Décodage} : ce module contient des scripts et des excusables permettant de trouver la traduction la plus probable d’une phrase source en consultant les modèles du module Train. 
\end{itemize}

\subsubsection{IRSTLM}
IRSTLM \citep{federico2007efficient} est une boite à outils utilisée pour la construction des  modèles de langage statistiques. L’avantage de cette boite à outils est de réduire les besoins de stockage ainsi que la mémoire lors de décodage. Par conséquent, cet outil nous permet de gagner du temps pour le chargement du modèle de langage. 

\subsubsection{BLEU:Métrique d'évaluation automatique }

Le score BLEU (en anglais : Bilingual Evaluation Understudy) a initialement été proposé par \cite{papineni2002bleu}.C’est un algorithme utilisé en vue d’évaluer la qualité des hypothèses de sortie produites par un système de traduction automatique. 

En effet, le concept est fondé sur l'idée de comparer l’hypothèse de traduction avec une ou plusieurs références au niveau des mots, des bigrammes, trigrammes etc.

 Le score BLEU est normalisé entre 0 et 1, et il est exprimé généralement en pourcentage. Notons qu’une traduction humaine peut parfois obtenir un mauvais score BLEU , si elle s’écarte de la référence.

\subsubsection{MADAMIRA}
L'analyseur morphologique MADAMIRA \cite{pasha2014madamira}  : est un système d'analyse morphologique et de désambiguïsation  de l'arabe qui exploite certains des meilleurs aspects des deux systèmes existants et les plus utilisés pour le traitement automatique de la langue arabe que sont : MADA (\cite{habash2005arabic}; \cite{habash2009mada+};. \cite{habash2013morphological}) et AMIRA \cite{diab2009second}.  En effet, MADAMIRA permet la tokenisation, la lemmatisation, le racinisation, l'étiquetage morpho-syntaxique, la désambiguïsation morphologique, la diacritisation, la reconnaissance des entités nommées, etc.

MADAMIRA propose les deux shémas de tokenisation suivants: 
\begin{itemize}
\item \textbf{ATB:} consiste à segmenter touts les clitiques excepté les articles définis, de même elle consiste à normaliser les caractères ALIF et YA en utilisant le caractère '+' comme étant un marqueur de clitiques.
\item \textbf{MyD3:}  consiste à tokeniser les  proclitiques QUES, CONJ, les clitiques PART, ainsi que touts les articles et enclitiques. En outre, elle normalise les caractères ALIF et YA  après la dévoyelisation des caractères arabes. 
\end{itemize}

\subsection{Corpus parallèles  }

\subsubsection{LDC-Ummah}
 Ummah (LDC2004T18) est un corpus de news historique arabe aligné avec des traductions  Anglais collectées via le service de presse \textit{Ummah} de Janvier 2001 à Septembre 2004. 
 
 Il totalise 8.439 paires histoire, 68,685 paires de phrases, de mots arabes et 2M mots 2,5M anglais. 
 Le corpus est aligné au niveau de la phrase. Tous les fichiers de données sont des documents SGML. 
 
 \begin{table}[!h]
 \centering
 \begin{tabular}{lccc}
  \hline & Nombre de mots & Nombre de lignes 
 \\\hline arabe & 2M   & 68,6 K 
 \\\hline Anglais & 2,4M & 68,6  K
 \\\hline
   
 \end{tabular}
 \caption{ Description des corpus Ummah }
 \end{table}
 
\subsubsection{LDC-News}
 le corpus LDC-News (Arabic News Translation Text Part 1)  a été produit par \textit{LDC} (Linguistic Data Consortium) sous le numéro de catalogue LDC2004T17.
  Trois sources de texte journalistique arabe ont été sélectionnés pour produire ce corpus arabe
 \begin{itemize}
 \item Service des nouvelles \textit{AFP}:  250 nouvelles, 44 193 mots arabes, octobre 1998 - décembre 1998 - \item  Service des ouvelles \textit{Xinhua}: 670 nouvelles histoires, 99 514 mots arabes, Novembre 2001 - Mars 2002 
 \item An Nahar : 606 nouvelles, 297 533 mots arabes, de Octobre 2001 - Décembre 2002 
  \begin{table}[!h]
\centering
 \begin{tabular}{lccc}
  \hline & Nombre de mots & Nombre de lignes 
 \\\hline arabe & 441 K &18,6 K
 \\\hline Anglais & 581 K & 18,6 K
 \\\hline
   
 \end{tabular}
 \caption{Description des corpus LDC-News }
  \end{table}
 \end{itemize}

\subsubsection{News Commentary}
 Le corpus News commentary  est un corpus parallèle
aligné au niveau des phrases. Ce corpus contient des extraits de diverses publications de presse et de commentaires du projet\textit{ Syndicate} et il est disponible dans plusieurs langues (arabe, anglais,  français, espagnol, allemand, et tchèque, etc).
  \begin{table}[!h]
   \centering
 \begin{tabular}{lccc}
  \hline & Nombre de mots & Nombre de lignes 
 \\\hline  arabe &3,9 M & 174,5 K 
 \\\hline  Anglais & 4,1 M & 174,5 K 
 \\\hline
   
 \end{tabular}
 \caption{ Description du corpus News Commentary }
 \end{table}
\subsubsection{TED Talks}
TED Talks est un ensemble de transcriptions des conférences en anglais présentés sous format vidéo sur le site officiel de TED. Ces transcriptions ont été traduites par les bénévoles pour plus de 70 autres langues (arabe, français, italien, coréen, portugais, etc.).

 \begin{table}[!h]
 \centering
 \begin{tabular}{lccc}
  \hline & Nombre de mots & Nombre de lignes 
 \\\hline arabe & 416 K  & 29,7 K 
 \\\hline Anglais &501 K &  29,7 K 
 \\\hline   
 \end{tabular}
 \caption{ Description du corpus TED }
 \end{table}
 
\section{Mise en place du système de TA anglais-arabe}

En arabe nous trouvons plusieurs clitiques qui se collent au mot, conduisant à des ambiguïtés morphologiques et orthographiques. Ainsi, pour construire un système de traduction Anglais-arabe, il est nécessaire de passer par une étape de segmentation du corpus au niveau des mots en pré-traitement (avant de construire le système de traduction) ainsi qu’une étape de détokenisation en post-traitement (après la traduction d’un corpus tokenisé). De ce fait, il est important de trouver le bon schéma de tokenisation à suivre qui ne se trompe pas en détectant le token et les clitiques, et de réussir à retourner après le format initial au texte arabe traduit. Diverses approches ont été proposées pour faire face aux problèmes (d’ambiguïté morphologique en arabe) de tokenisation et détokenisation en arabe. Dans l’un des premiers ouvrages, et d’ailleurs l’un des plus connus dans ce domaine \cite{habash2006arabic} ont présenté différents schémas de tokenisation pour le pré-traitement de l’arabe en vue de voir quelle est la méthode de segmentation la plus utile pour la TAS. Ces schémas sont disponibles dans l’outil MADAMIRA que nous avons utilisé.
Nous avons construit un système de traduction automatique statistique à l'aide de la boite à outils Moses ainsi que IRSTLM pour créer notre modèle de langage 5-grammes, et en utilisant les corpus parallèles décrits précédemment (LDC-Ummah, LDC-News, News Commentary, TED Talks). Nous avons évalué notre système en termes du score BLEU (score de 24,51).
 
\section{Conclusion et Perspectives}

Dans cet article, nous avons présenté notre système de traduction anglais-arabe basé sur la boite à outils Moses, construit à l'aide d'un modèle de langage 5-grammes et en utilisant différents corpus parallèles que nous avons décrits. Nous envisageons d'exploiter notre système pour traduire de grands corpus de l'anglais vers l'arabe.


~ \newpage

\bibliographystyle{taln2018.bst}
\bibliography{biblio}

\begin{thebibliography}{~~~}

\bibitem[\protect\citename{Diab, }2009]{diab2009second}
{\sc Diab M.} (2009).
\newblock Second generation amira tools for arabic processing: Fast and robust
  tokenization, pos tagging, and base phrase chunking.
\newblock In {\em 2nd International Conference on Arabic Language Resources and
  Tools}.

\bibitem[\protect\citename{Federico \& Cettolo, }2007]{federico2007efficient}
{\sc Federico M. \& Cettolo M.} (2007).
\newblock Efficient handling of n-gram language models for statistical machine
  translation.
\newblock In {\em Proceedings of the Second Workshop on Statistical Machine
  Translation}, p.\ 88--95: Association for Computational Linguistics.

\bibitem[\protect\citename{Habash \& Rambow, }2005]{habash2005arabic}
{\sc Habash N. \& Rambow O.} (2005).
\newblock Arabic tokenization, part-of-speech tagging and morphological
  disambiguation in one fell swoop.
\newblock In {\em Proceedings of the 43rd Annual Meeting on Association for
  Computational Linguistics}, p.\ 573--580: Association for Computational
  Linguistics.

\bibitem[\protect\citename{Habash {\em et~al.}, }2009]{habash2009mada+}
{\sc Habash N., Rambow O. \& Roth R.} (2009).
\newblock Mada+ tokan: A toolkit for arabic tokenization, diacritization,
  morphological disambiguation, pos tagging, stemming and lemmatization.
\newblock In {\em Proceedings of the 2nd international conference on Arabic
  language resources and tools (MEDAR), Cairo, Egypt}, p.\ 102--109.

\bibitem[\protect\citename{Habash {\em et~al.}, }2013]{habash2013morphological}
{\sc Habash N., Roth R., Rambow O., Eskander R. \& Tomeh N.} (2013).
\newblock Morphological analysis and disambiguation for dialectal arabic.
\newblock In {\em Hlt-Naacl}, p.\ 426--432.

\bibitem[\protect\citename{Habash \& Sadat, }2006]{habash2006arabic}
{\sc Habash N. \& Sadat F.} (2006).
\newblock Arabic preprocessing schemes for statistical machine translation.
\newblock In {\em Proceedings of the Human Language Technology Conference of
  the NAACL, Companion Volume: Short Papers}, p.\ 49--52: Association for
  Computational Linguistics.

\bibitem[\protect\citename{Koehn {\em et~al.}, }2007]{koehn2007moses}
{\sc Koehn P., Hoang H., Birch A., Callison-Burch C., Federico M., Bertoldi N.,
  Cowan B., Shen W., Moran C., Zens R. {\em et~al.}} (2007).
\newblock Moses: Open source toolkit for statistical machine translation.
\newblock In {\em Proceedings of the 45th annual meeting of the ACL on
  interactive poster and demonstration sessions}, p.\ 177--180: Association for
  Computational Linguistics.

\bibitem[\protect\citename{Kuhn \& De~Mori, }1990]{kuhn1990cache}
{\sc Kuhn R. \& De~Mori R.} (1990).
\newblock A cache-based natural language model for speech recognition.
\newblock {\em IEEE transactions on pattern analysis and machine intelligence},
  {\bf 12}(6), 570--583.

\bibitem[\protect\citename{Lau {\em et~al.}, }1993]{lau1993trigger}
{\sc Lau R., Rosenfeld R. \& Roukos S.} (1993).
\newblock Trigger-based language models: A maximum entropy approach.
\newblock In {\em Acoustics, Speech, and Signal Processing, 1993. ICASSP-93.,
  1993 IEEE International Conference on}, volume~2, p.\ 45--48: IEEE.

\bibitem[\protect\citename{Och \& Ney, }2003]{och2003systematic}
{\sc Och F.~J. \& Ney H.} (2003).
\newblock A systematic comparison of various statistical alignment models.
\newblock {\em Computational linguistics}, {\bf 29}(1), 19--51.

\bibitem[\protect\citename{Papineni {\em et~al.}, }2002]{papineni2002bleu}
{\sc Papineni K., Roukos S., Ward T. \& Zhu W.-J.} (2002).
\newblock Bleu: a method for automatic evaluation of machine translation.
\newblock In {\em Proceedings of the 40th annual meeting on association for
  computational linguistics}, p.\ 311--318: Association for Computational
  Linguistics.

\bibitem[\protect\citename{Pasha {\em et~al.}, }2014]{pasha2014madamira}
{\sc Pasha A., Al-Badrashiny M., Diab M.~T., El~Kholy A., Eskander R., Habash
  N., Pooleery M., Rambow O. \& Roth R.} (2014).
\newblock Madamira: A fast, comprehensive tool for morphological analysis and
  disambiguation of arabic.
\newblock In {\em LREC}, volume~14, p.\ 1094--1101.

\end{thebibliography}

\end{document}